\documentclass{article}

     \PassOptionsToPackage{numbers, compress}{natbib}

\usepackage[preprint]{neurips_2024}

\usepackage[utf8]{inputenc} 
\usepackage[T1]{fontenc}    
\usepackage{hyperref}       
\usepackage{url}            
\usepackage{booktabs}       
\usepackage{amsfonts}       
\usepackage{nicefrac}       
\usepackage{microtype}      
\usepackage{xcolor}         
\usepackage{graphicx} 
\usepackage{amsmath} 

\title{Conversational Planning for Personal Plans}

\author{%
  Konstantina Christakopoulou  \\
  Google DeepMind
  \and \textbf{Iris Qu} \\
  Google DeepMind
  \and \textbf{John Canny} \\
  Google DeepMind
  \AND 
  \textbf{Andrew Goodridge} \\
  Google
  \and \textbf{Cj Adams} \\
  Google 
  \and \textbf{Minmin Chen} \\
  Google DeepMind 
  \and \textbf{Maja Matari\'c} \\
  Google DeepMind
}

\begin{document}

\maketitle

\begin{abstract}
The language generation and reasoning capabilities of large language models (LLMs) have enabled conversational systems with impressive performance in a variety of tasks, from code generation, to composing essays, to passing STEM and legal exams, to a new paradigm for knowledge search. Besides those short-term use applications, LLMs are increasingly used to help with real-life goals or tasks that take a long time to complete, involving multiple sessions across days, weeks, months, or even years. Thus to enable conversational systems for long term interactions and tasks, we need language-based agents that can plan for long horizons. Traditionally, such capabilities were addressed by reinforcement learning agents with hierarchical planning capabilities. In this work, we explore a novel architecture where the LLM acts as the meta-controller deciding the agent's next macro-action, and tool use augmented LLM-based option policies execute the selected macro-action. We instantiate this framework for a specific set of macro-actions enabling adaptive planning for users' personal plans through conversation and follow-up questions collecting user feedback. We show how this paradigm can be applicable in scenarios ranging from tutoring for academic and non-academic tasks to conversational coaching for personal health plans.

\end{abstract}

\section{Introduction}
\label{sec:intro}

Consider would happens when we ask another person about a complex real-life task, such as planning a vacation,  learning how to draw, or improve one's fitness level. If the person does not know you, they would first ask you a couple of questions to clarify your preferences, your skill level, and your goals. Having gained a better understanding about you and the task, they might provide some suggestions for you to consider, and then keep adapting those as you provide more feedback through conversation. For example, for the vacation planning task, they might start with steps such as "determine the budget" and "figure out the preferred kind of vacation", and "narrow down potential destinations". For each step, they might ask clarifying questions, or answer your questions and provide some relevant resources like a web site for a relevant city or airline. As you start getting more concrete about the specifics of your goal, and take actions such as choosing the destination, you need help with the next step, such as booking accommodations and planning specific activities.  Therefore, the plan evolves---either the person recommends new steps, revises some existing steps to account for the updated information, or asks further questions. A similar \emph{conversational planning} process is followed whenever a fitness coach is asked to provide a plan adapted for their client, and whenever a tutor needs to provide a personalized learning plan.  All of these plans need to evolve over time based on the updated needs of the user.

However, when we ask the popular conversational interfaces of today to help with long term real-life tasks or goals, they behave very differently than people. They tend to generate long responses with bulleted lists of steps to follow. They do not ask questions proactively, but only respond in a reactive way to user queries. Also, they do not provide a structured plan for the user to follow that can be adapted over time based on user feedback. In this work, we aim to bridge this gap between how conversational agents and humans solve long-term tasks and goals.

We argue that, thanks to the impressive reasoning and natural language generation capabilities of large language models (LLMs), doing interactive hierarchical planning via natural language is an effective approach for artificial intelligence (AI) agents to assist and collaborate with users about their plans for real-life goals and tasks. We propose a framework where a chain-of-thought (CoT)-prompted LLM \cite{wei2022chain} plays the role of the meta-controller deciding the next macro-action, which is then executed by options/policies powered via the same or different LLM, CoT-prompted with different examples. The meta-controller receives natural language feedback from the user, allowing for interactive planning over time. This general LLM-based hierarchical framework can be used for various long-term planning tasks---we instantiate it in the context of assisting a user with a long-term real-life goal. Thus, the macro-actions are instantiated to "ask a question", "add a step", or "alter a step", so that the AI agent can operate in a similar fashion as a person would if they were to be asked to help with coaching or tutoring or assisting with making a plan. For each step of the plan, as executed by the LLM-based policy, the system also fetches and ranks content via tool use and LLM CoT prompting \cite{yao2022react}, so that the agent can point to specific resources the user might want to look at to further explore that step of the plan. The plan can be adapted over time via  natural language user feedback, obtained either via user's answers to the questions or via user's free-form feedback. We qualitatively demonstrate the effectiveness of the approach across domains of learning and health.

\section{Related Work}


Inspired by the strong in-context learning \cite{brown2020language, kojima2022large} and complex reasoning \cite{wei2022chain, zhou2024self} capabilities of LLMs \cite{bubeck2023sparks}, there has been a lot of recent interest into \textbf{LLM-driven agents} \cite{wang2023describe, zhao2024empowering, ahn2022can, shen2024hugginggpt, nakano2021webgpt, wang2023voyager, wang2024survey, sharan2023llm, brooks2024large}. This body of work can be largely divided into (i) text-based agents \cite{shen2024hugginggpt, yao2022react, park2023generative}, and (ii) open-world embodied agents \cite{wang2023describe, ahn2022can, wang2023voyager, liang2023code, singh2023progprompt, szot2023large, sun2024interactive}. Our work, anchored on the realization that real-life user goals/tasks take a long time to be achieved, brings language-based \textbf{hierarchical long-term planning} techniques typically considered only in open-world environments (e.g., Minecraft) to text-based agents. Traditionally, in non-language based reinforcement learning, whenever tasks of long term nature need to be solved, a \emph{hierarchical} goal execution architecture is considered \cite{bacon2017option}, where a planner generates action plans that would then be executed by low-level goal-conditioned controllers. Recently, thanks to the LLM advances, interactive LLM-based hierarchical planning approaches have been proposed and proven successful for open-world environments \cite{sun2024interactive, singh2023progprompt, wang2023voyager, huang2022language, zhao2024empowering, wang2023describe}. However, little to no attention has been given to language-based hierarchical planning for text-based agents \cite{shen2024hugginggpt}. This is where our work comes in and aims to advance the frontier of the traditionally short-term focused text-based agents. We offer a novel framework allowing for conversational planning for users' personal plans, that can be adapted over time based on human feedback. This also advances the state of the art for conversational recommendation \cite{christakopoulou2016towards, jannach2021survey}.



\section{Proposed Approach}
\begin{figure}[!t]
    \centering
    \includegraphics[scale=0.13]{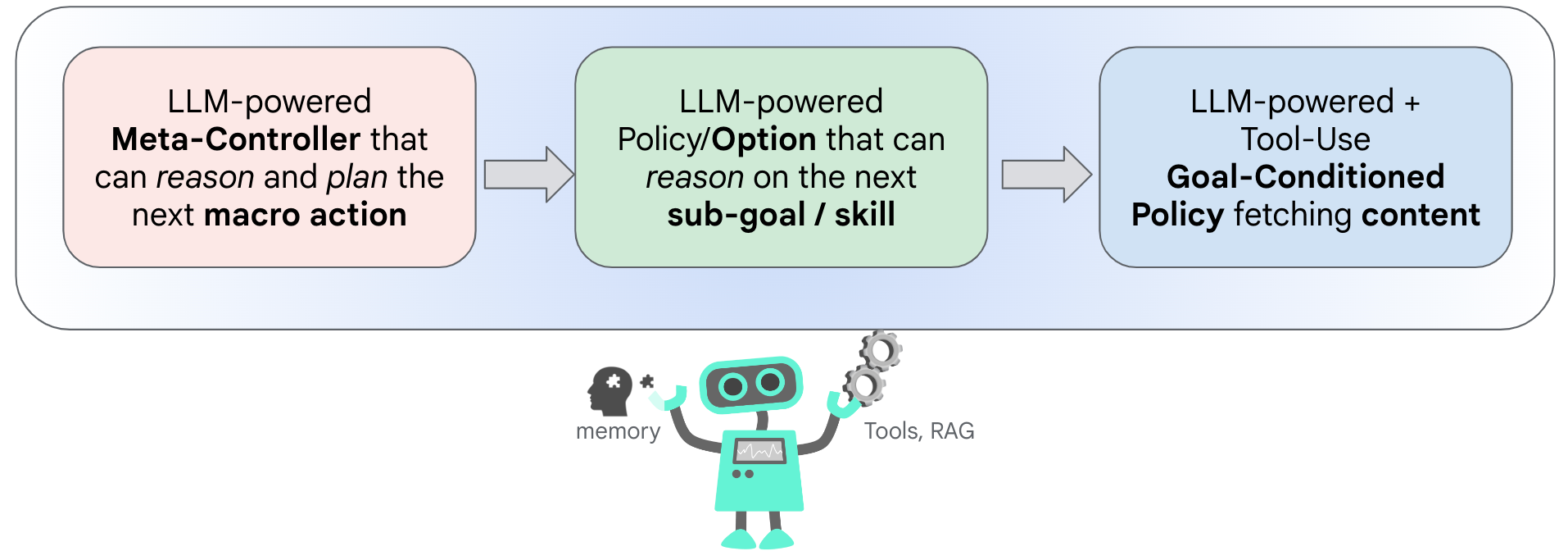}
    \caption{Overview of hierarchical planning via natural language.}
    \label{fig:overview}
\end{figure}

As described in Section \ref{sec:intro}, the goals and tasks users pursue can last for multiple days, months, or even years \cite{christakopoulou2023large}. Developing AI agents assisting towards these can be framed as a \emph{long-horizon} planning problem \cite{sutton1999reinforcement}. Traditionally, this would be framed in a hierarchical reinforcement learning (RL) framework, where the underlying hierarchy of options/ skills would need to be given or learned \cite{bacon2017option}. To enable \emph{conversational} AI agents that can help along the user's journey towards these goals/ tasks in a similar way a human would, and that can adapt based on user feedback, we need to address the following research question:
\emph{How might we do \textbf{hierarchical planning via natural language} to assist with the users' \textbf{plans} serving their real-life goals / tasks?}

\subsection{Overview: Language-based Interactive Hierarchical Planning}
\noindent \textbf{RL Framing.} We frame the conversational AI system as a language-based agent that can interact with the user $u$ who is part of the world/environment $\mathcal{E}$. The user starts by using language to specify the goal they want to pursue $g \in \mathcal{L}$ --- this can be viewed as the initial observation $o_1$.  The agent can interact with the user only via natural language feedback $o \in \mathcal{L}$. The feedback can contain new observations about the user, allowing the user to provide more information about themselves and their needs/goals/preferences/how their journey towards this goal is evolving (e.g., \emph{"I've already chosen my vacation destination"}/skill level (e.g., \emph{"I have prior drawing experience"} for a learning journey), and rewards about the provided plan (e.g., \emph{"I don't like this plan"}, or \emph{"Can you add more steps on relaxation techniques before going to bed?"} for a health journey). Given the observations/feedback $o$ from the user $u$, the agent decides the next action $a$ to take, which changes the state from $s_t$ to $s_{t+1}$. The agent can only see partial observations about the user's state based on the natural language feedback. This can be framed as a POMDP \cite{sutton1999reinforcement}, where the agent takes actions according to a $\theta$-parameterized policy, i.e., $a \sim \pi( a | s; \theta)$. We consider language-based agents with policies instantiated via CoT-prompted LLMs.

\noindent \textbf{Hierarchical RL Framing.} The agent can take various actions, such as choosing content resources, asking a question, and providing a plan for the user to follow. Analogously to human reasoning, the agent can make these decisions in a hierarchical fashion, starting from a high level and proceeding to more granular actions. This allows for dealing with long horizons. At every user-agent interaction, rather than providing a single (primitive) action, the agent needs to choose a macro-action / goal / option / skill at every $c$ steps. Then given that macro-action, more granular primitive actions can be selected. This corresponds to a hierarchical policy, consisting of a \emph{high-level policy} or \emph{meta-controller} deciding the macro-action (Section \ref{subsec:meta-controller}), and the \emph{low-level policy}, a sub-policy over primitive actions (Sections \ref{subsec:low-level-policy}, \ref{subsec:tool-use}), both updated over time based on user feedback and with potentially different temporal granularity. This paradigm has been long considered in the RL literature \cite{sutton1999reinforcement, bacon2017option}, but has received little attention in the context of LLMs assisting long-term user goals/tasks. Next, we detail the language-based meta-controller and sub-policies. The overall framework is shown in Figure \ref{fig:overview}.
  
\subsection{Proposed Framework Components}

\noindent \textbf{Language-based Meta-Controller: Deciding to ask questions, add or alter plan steps}
\label{subsec:meta-controller}
The meta-policy is instantiated via a CoT-prompted LLM \cite{wei2022chain} capable of generating structured language objects such as JSON \cite{zhou2024self}. The space of actions $\mathcal{Z}$ is discrete, i.e., \texttt{add-steps}, \texttt{alter-steps}, and \texttt{ask-question}, but the underlying LLM can also generate natural language arguments for each macro-action $x_z$, e.g., the specific step names to add or alter, or more context about the question asked. 
The LLM can generate an action $z$, generate arguments for that action, and reason about its actions through thoughts $\tau \in \mathcal{L}$. The thoughts, actions, and arguments for the actions become the structured language meta-actions the controller can take, denoted as $\hat{z}$, with $\hat{\mathcal{Z}} = \mathcal{Z} \cup \mathcal{T} \cup \mathcal{X} \subseteq \mathcal{L}$. Thus, if $\Phi$ denotes all the parameters of the LLM implementing the policy, then the meta-controller selects the macro-action $\hat{z}$ given the context $c_t$, which includes the user's initial goal $g$, user feedback $o_t$, the history of user-system interactions $H$, and previous macro-actions the meta-controller has taken $\hat{z}_0, \ldots, \hat{z}_{t-1}$: 
$\text{Meta-Controller: ~~}\pi(\hat{z} | c_t; \Phi), \text{ where context } c_t = \text{Concat}(o_t, \mathcal{H}, \hat{z}_0, \ldots, \hat{z}_{t-1}, g).\\
$

\noindent \textbf{Language-based sub-policies: Generating the added/altered steps or question}
\label{subsec:low-level-policy}
Once the meta-controller selects the macro-action $\hat{z}$, the corresponding language-based sub-policy comes into play. The sub-policies are implemented also via (potentially the same) LLM, but CoT-prompted with different examples. We have three different language sub-policies, one per discrete action $z$ of the meta-controller: $\pi_{\texttt{add-steps}}(\cdot | ; \Omega_1)$, $\pi_{\texttt{alter-steps}} (\cdot | ; \Omega_2)$ and $\pi_{\texttt{ask-question}} (\cdot | ; \Omega_3)$, each with parameters $\Omega_1, \Omega_2$, and $\Omega_3$, respectively. The $\pi_{\texttt{add-steps}}$ generates sets of actions in the structured language space. After instantiating the specific schema each plan step should follow (name, description, follow-up per step, search keywords used by tool-use policies), the LLM-based policy for adding steps generates a tentative plan of new steps along with the reasoning/thought behind it, as well as a summarization of what it has learned about the user so far. Similarly, the $\pi_{\texttt{alter-steps}}$, CoT-prompted with different examples illustrating how a step should be altered, maps the existing plan along with the name of the step to be altered as generated by the argument of the meta-controller, to an altered step, again living in the same structured language schema described above. The $\pi_{\texttt{ask-question}}$ policy maps the meta-controller macro-action to the specific question to ask the user. Note that when the sub-policy asks a question, the plan remains unchanged. These questions are different from the questions the $\pi_\texttt{add-steps}$ and $\pi_{\texttt{alter-step}}$ generate per plan step, which the system collects across the steps of the plan in order to enable the user to choose one as a way to continue the planning process.\\

\begin{figure*}[t]
    \centering
    \includegraphics[scale=0.14]{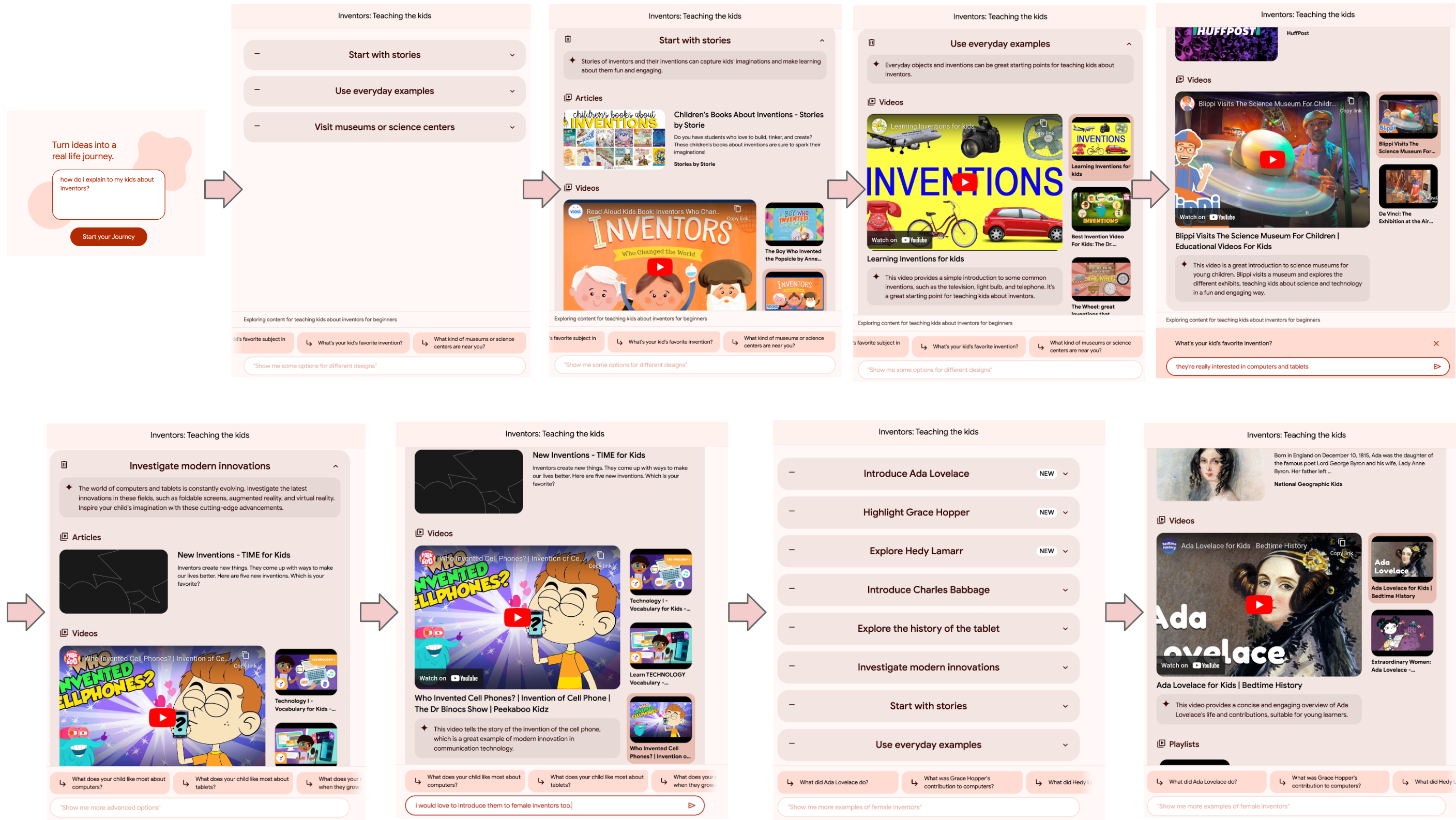}
    \caption{Demonstration of "explaining to my kids about inventors".}
    \label{fig:inventors_flow}
\end{figure*}
\noindent \textbf{Language-based Low-Level Policies: Connecting the user with content per plan step}
\label{subsec:tool-use}
Framing recommendation as an RL problem, the action the system takes at every step is the content item $c$ to show to the user \cite{chen2019top}. This becomes the primitive action space of our language-based hierarchical agent. The language-based low-level policies decide what content to show to the user per plan step via a combination of \emph{tool use} and \emph{CoT-prompted LLMs}. Specifically, for each step in the updated plan, after the execution of the \texttt{add-steps} or \texttt{alter-step} policy, the LLM can decide which tool to call (e.g., \texttt{SEARCH}, \texttt{RECOMMEND-ENGINE}) and fetch $n$ content items per step. Then, given these fetched content items, the same, but different CoT-prompted LLM can decide which of those items should be in the top-$k$ that are shown to the user. In other words, we approximate the retrieval-ranking two-stage recommendation approach \cite{covington2016deep} by tool use and CoT prompting, respectively.\\

\noindent \textbf{Interactive Planning based on User's Natural Language Feedback}
Using LLMs as the meta-controller, the sub-policies, and the low-level policies, the AI agent can create or update a plan, and ask a question to assist with a user's goal $g \in \mathcal{G} \subset \mathcal{L}$ expressed in natural language. As the user's journey towards this goal evolves over time, the user can provide natural language feedback $o$ which is then used to update both the meta-controller in terms of the next macro-action to take, the sub-policies in terms of the specific steps to add / alter or specific questions to ask, and the specific content items to include per plan step. The policies are updated by including user feedback in the context window, as is done with natural language-based RL agents \cite{shinn2024reflexion, zhao2024empowering}. This mechanism allows for the plan to be interactively adapted and personalized to different users.

\section{Qualitative Evaluation}
\label{sec:eval}

\begin{figure*}[!t]
    \centering
    \includegraphics[scale=0.07]{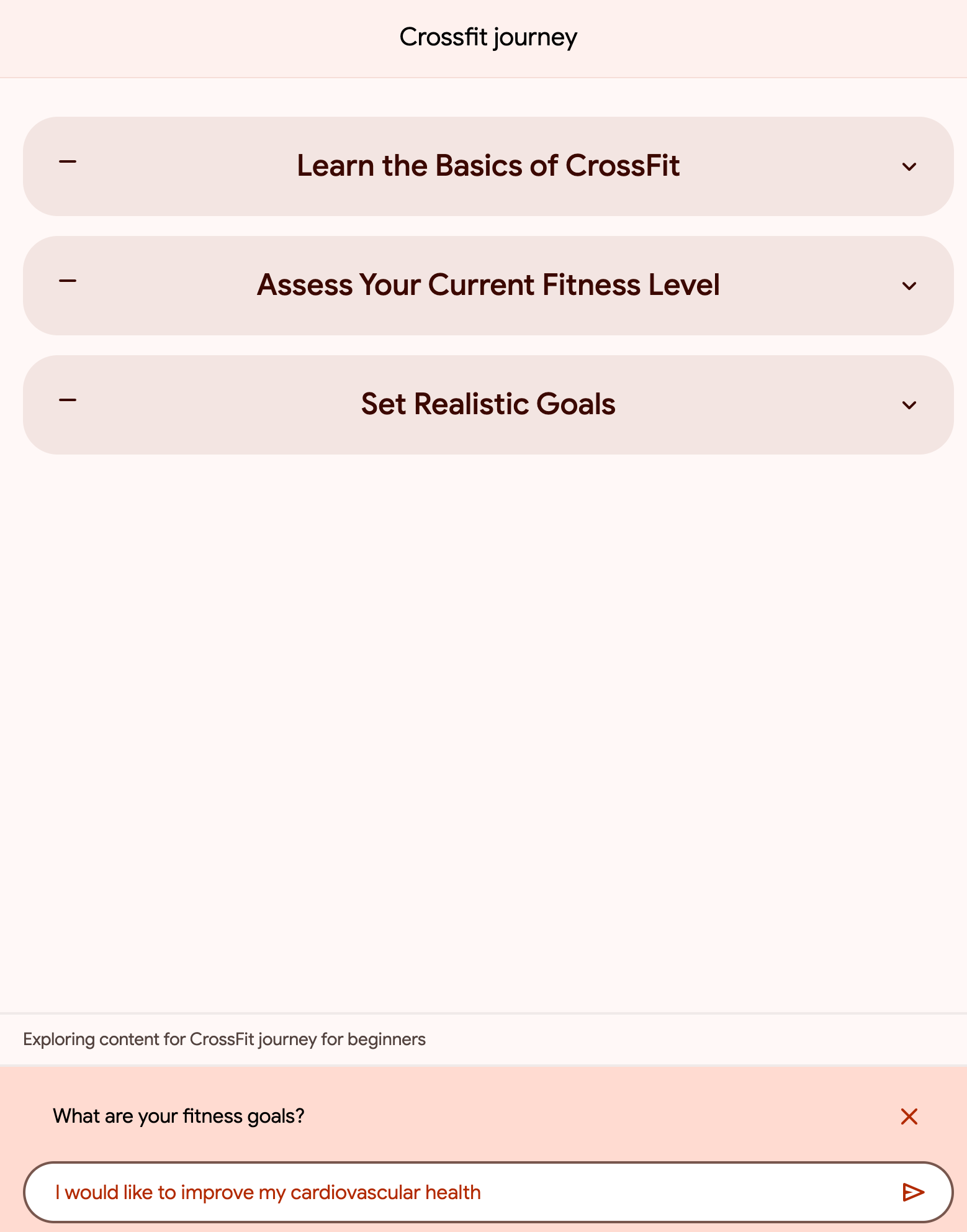}
    \includegraphics[scale=0.07]{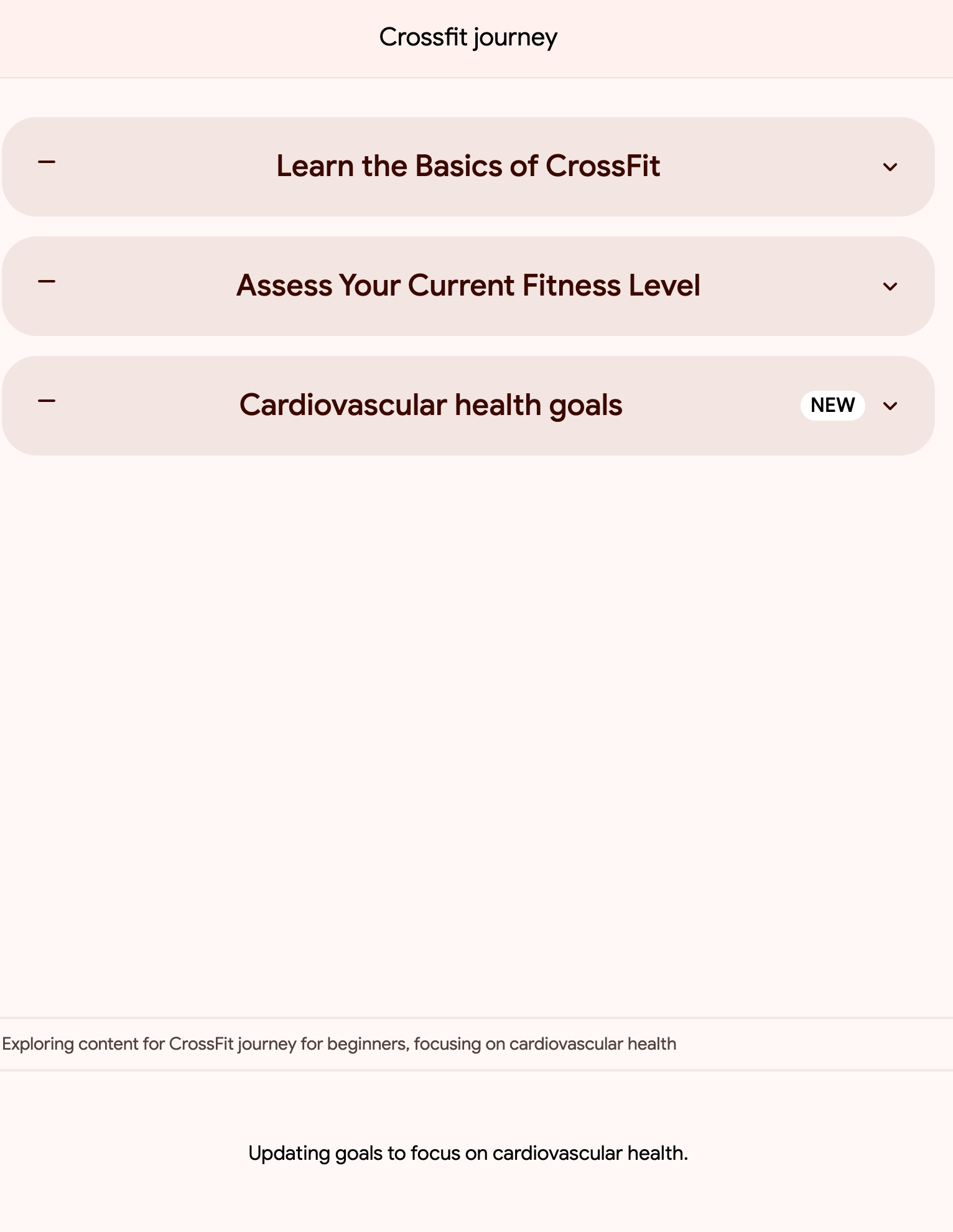}
    \includegraphics[scale=0.07]{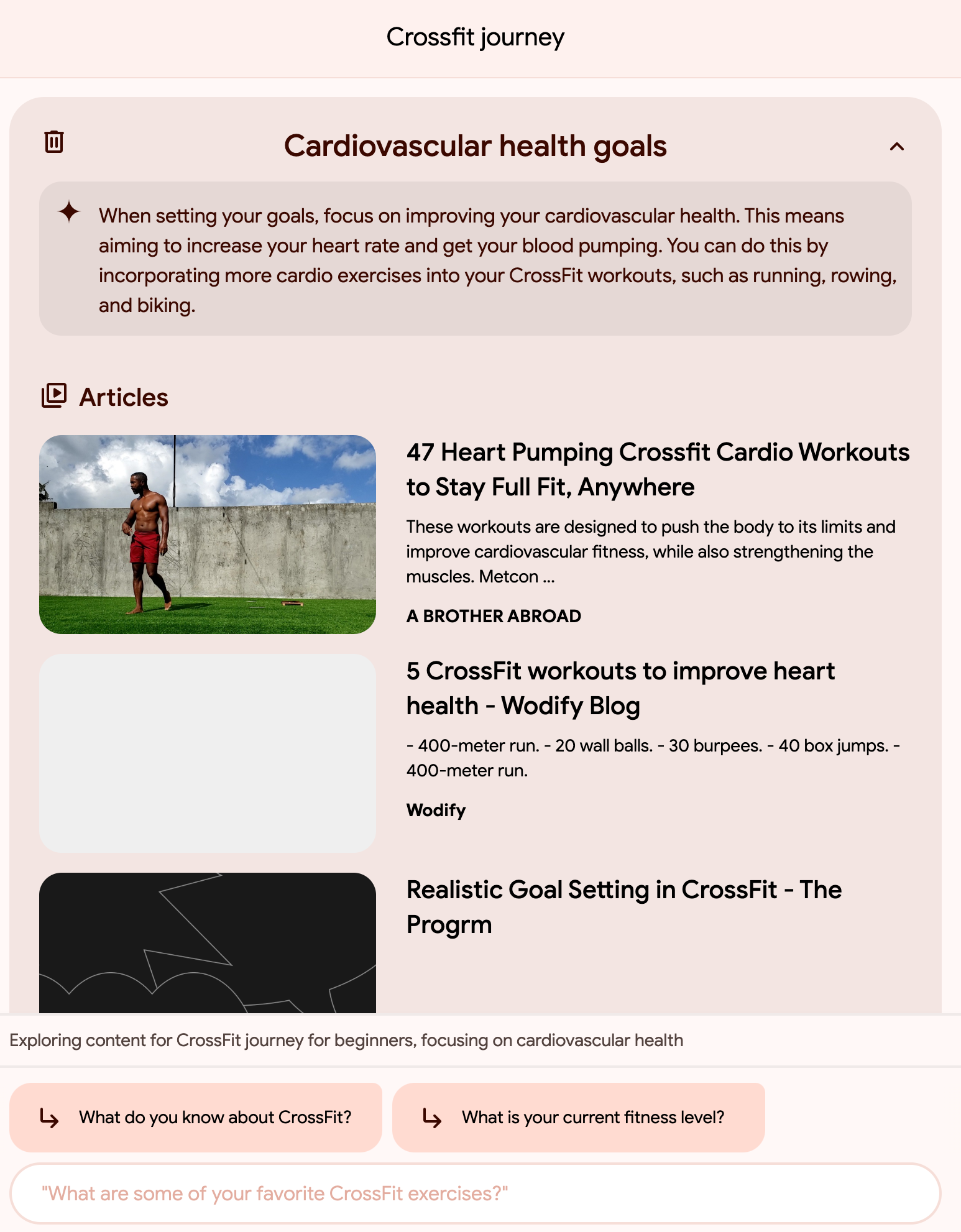}
    \caption{Conversational Coaching for crossfit fitness journey}
    \label{fig:fitness_journey}
\end{figure*}
We performed a qualitative evaluation and found that our proposed framework is powerful in providing interactive planning for user goals across a range of domains, including learning skills, pursuing DIY projects, hobbies, fitness, health, aspirational goals, and many more nuanced personal journeys. Here, we show examples from two important real-world domains: learning \cite{kizilcec2020scaling} and health \cite{davenport2019potential}.  

\noindent \textbf{Conversational Tutoring for a Learning Goal}
Figure \ref{fig:inventors_flow} demonstrates an example of a user-agent interaction in which the user asks for a learning plan: \emph{"How do I explain to my kids about inventors?"} We can see that given the initial query, the meta-controller decided to add three new steps with names \emph{Start with stories}, \emph{Use everyday examples}, and \emph{Visit museums or science centers}. Each of these steps is structured, as executed by sub-policy of \texttt{add-steps}, generating the aforementioned names, a description of why this step is useful, and search keywords. The search keywords are then used by the low-level tool-use policies to fetch and rank content from \texttt{SEARCH}, and \texttt{RECOMMEND-ENGINE}. The content includes animation-based videos and articles regarding children's books on inventions. Additionally, the sub-policy generated follow-up questions per each plan step, as shown at the bottom. The user can interact via either answering one of the provided questions (e.g., Q: \emph{"What's your kids favorite invention?"}, A: \emph{"They're really interested in computers and tablets"}), or by writing their own natural language feedback, i.e., \emph{"I would love to introduce them to female inventors too."}. This feedback becomes part of the context of the agent across all levels of the hierarchical policy, and leads to a plan adaptation where new steps on \emph{Investigate modern inventions}, \emph{Explore the history of the tablet}, \emph{Introduce Charles Babbage} are added to the plan. The plan is further adapted based on the user's free-form feedback on female inventors to include steps on \emph{Ada Lovelace}, \emph{Grace Hopper}, and \emph{Hady Lamarr}. The figure illustrates that the LLM-based policies  fetch relevant content and generate appropriate follow-up questions per step to allow the user to propel their journey forward.  

\noindent \textbf{Conversational Coaching for Personal Health}
Similarly, Figure \ref{fig:fitness_journey} demonstrates an example of a user-agent interaction regarding a personal health journey in the domain of fitness---\emph{"I want to do crossfit"}. The hierarchical agent again  initially decided to \texttt{add-steps}, which then the CoT-prompted LLM-based sub-policy mapped to three steps in the plan: \emph{Learn the basics of crossfit}, \emph{Assess your current fitness level}, and \emph{Set realistic goals}. The tool-enhanced low-level policy then fetched and ranked content for each of the steps. The user can  interact with the sub-policy generated follow-up questions or write their own free-form feedback to further adapt the plan. Here, the user answered the question \emph{"What are your fitness goals?"}, by saying \emph{"I would like to improve my cardiovascular health."} Then, the meta-controller decided to \texttt{alter-step} with argument \emph{Set Realistic goals}. Given the meta-controller's decision, the corresponding sub-policy $\pi_{\texttt{alter-step}}$ executed the macro-action and altered the step, and the low-level policy used the updated search keywords to update the step.

\section{Conclusions}
\label{sec:concl}
We proposed a language-based hierarchical agent that is able to assist users across their real-life journeys towards goals and tasks that can last across sessions and evolve over long periods of time. The framework is general and aims to behave in a similar fashion as a human when asked to assist with a plan for a real-life goal. We demonstrate qualitatively the effectiveness of the framework for conversational coaching and conversational tutoring, noting success across other domains as well. We argue that this approach makes a step toward conversational AI systems that are not anchored in short-term interactions but can be true companions in supporting user goals over time.

\bibliographystyle{ACM-Reference-Format}
\bibliography{neurips_2024}

\end{document}